\documentclass[supp]{new_tlp}

\usepackage{supp} 
\usepackage{times}
\usepackage{helvet}
\usepackage{courier}
\usepackage{xspace}
\usepackage[usenames,dvipsnames]{color}
\usepackage{amsmath}
\usepackage{url}
\usepackage[all]{xy}
\pubauthor{Aziz} 

\newcommand{\printsolver}[1]{\textsc{#1}\xspace}
\newcommand{\chuffed}{\printsolver{chuffed}}

\newcommand{\clingcon}{\printsolver{clingcon}}

\newcommand{\gringo}{\printsolver{gringo}}

\newcommand{\clasp}{\printsolver{clasp}}

\newcommand{\smodels}{\printsolver{smodels}}

\newcommand{\MiniZinc}{\printsolver{MiniZinc}}
\newcommand{\minizinc}{\MiniZinc}

\newcommand{\cmodels}{\printsolver{cmodels}}
\newcommand{\ezcsp}{\printsolver{ezcsp}}

\newcommand{\false}{\mathit{false}}
\newcommand{\true}{\mathit{true}}

\newcommand{\vars}{\mathit{vars}}

\newcommand{\mn}{\mathit{min}}
\newcommand{\mx}{\mathit{max}}

\newcommand{\rules}{\mathit{rules}}

\newcommand{\VV}{{\cal V}}

\newcommand{\AFV}{{\cal F}}
\newcommand{\ANV}{{\cal S}}

\DeclareMathSymbol{\naf}{\mathord}{symbols}{"18}

\newcommand{\ignore}[1]{}
\newcommand{\cupp}{\ensuremath{\cup}:=~}

\def\inc(#1,#2){#1_{#2}^{\uparrow}}
\def\dec(#1,#2){#1_{#2}^{\downarrow}}
\def\nonmon(#1,#2){#1_{#2}^{\updownarrow}}
\def\scope(#1,#2){#1|_#2}
\def\fdom(#1,#2,#3){#1_{#2}^{#3}}
\def\tpi(#1){T_P \uparrow {#1}}
\def\jb(#1,#2){\mathit{jb}_{#2}({#1})}
\def\push(#1,#2){{#1} \cupp {#2}}
\def\maxd(#1,#2){\mx(#1(#2))}
\def\mind(#1,#2){\mn(#1(#2))}

\newcommand{\minf}{-\infty}

\newcommand{\red}[2]{{#1}^{#2}}

\newtheorem{example}{Example}

\newcommand{\noproofs}[1]{}

\title{Bound Founded Answer Set Programming}
\author[R.A. Aziz]
	{Rehan Abdul Aziz \\
	Supervisors: Peter Stuckey and Geoffrey Chu \\
	National ICT Australia, Victoria Laboratory,\thanks{NICTA is funded by the Australian Government as represented by the
Department of Broadband, Communications and the Digital Economy and the
Australian Research Council through the ICT Centre of Excellence program.}
        \\
	Department of Computing and Information Systems \\
	Room 6.22, Doug McDonell Building (Building 168), \\
	The University of Melbourne, Australia \\
	Email:~{\tt raziz@student.unimelb.edu.au} \\
	}
\begin{document}

\label{firstpage}

\maketitle

\begin{abstract}
Answer Set Programming (ASP) is a powerful modelling formalism that is very efficient in solving combinatorial problems.
ASP solvers implement the stable model semantics that eliminates circular derivations between Boolean variables from the solutions
of a logic program. Due to this,
ASP solvers are better suited than propositional satisfiability (SAT) and Constraint Programming (CP) solvers to solve a certain class of problems whose specification includes inductive
definitions such as reachability in a graph. On the other hand, ASP solvers suffer from the \emph{grounding bottleneck} that occurs
due to their inability to model finite domain variables. Furthermore, the existing stable model semantics are not sufficient to disallow
circular reasoning on the bounds of numeric variables. An example where this is required is in modelling shortest paths between nodes in a graph.
Just as reachability can be encoded as an inductive definition with one or more base cases and recursive rules,
shortest paths between nodes can also be modelled with similar base cases and recursive rules for their \emph{upper bounds}. 
This deficiency of stable model semantics introduces another type of grounding bottleneck in ASP systems that 
cannot be removed by naively merging ASP with CP solvers, 
but requires a theoretical extension of the semantics from Booleans and normal rules to bounds over numeric
variables and more general rules. In this work, we propose Bound Founded Answer Set Programming (BFASP) that resolves this issue
and consequently, removes all types of grounding bottleneck inherent in ASP systems.
\end{abstract}

\section{Motivation}
Answer Set Programming \cite{baral03} is a useful modelling paradigm to solve search and planning problems.
Modern ASP solving \cite{clasp,clasp_journal} builds on propositional satisfiability (SAT) solving \cite{sat_solver}.
However, ASP solvers have a competitive edge over SAT solvers in problems 
whose model involves some notion of transitive closure, e.g., reachability or connectivity in a graph. 
This is due to the difference in semantics of both
systems; ASP solvers implement stable model semantics \cite{stable} which minimizes the number of variables that are true in a given
logic program while a SAT solver only looks for an assignment that satisfies all the given clauses. In ASP, in order for a
variable to be true, it must have some rule as a \emph{support} that \emph{justifies} it being true.
Furthermore, no set of variables can circularly support one another. 
E.g. given two rules $a \leftarrow b$ and $b \leftarrow a$, the only valid solution in stable model semantics
is where $a$ and $b$ are both false, whereas in propositional semantics, both the variables being true is also a valid
solution. 

As ASP systems such as \smodels and \clasp only deal with Boolean variables, they are inefficient for
solving problems that are naturally modelled with integers, especially if they have large domains. These combinatorial problems (e.g. scheduling)
are ubiquitous in Computer Science, which makes ASP a poor choice to model and solve them. The most obvious way to model these in ASP
is to represent each integer's domain as a set of Boolean variables and impose constraints
on these variables to ensure consistency. This incurs what is commonly known as the \emph{grounding bottleneck}
problem. Naturally, decomposing a large finite domain to Boolean variables blows up the problem size
in the size of domains of integer variables. Constraint Programming solvers \cite{cp_handbook} and
Mixed Integer Programming solvers, on the other hand, are excellent candidates for these problems as they support
numeric variables natively.
Unfortunately, constraint solvers suffer from the same inefficiency as SAT solvers regarding problems like
reachability that require inductive definitions. A hybrid system that has the best of both worlds, i.e., inductive rules for Boolean variables from ASP, 
and native support for integers and constraints over them from CP, 
addresses both the concerns. However, in this work, we propose that even this
hybrid approach is not sufficient, and there exists a type of grounding bottleneck that is still not removed by combining the strengths
of CP and ASP solvers.
Let us illustrate this by first looking at a benchmark from ASP competitions, and then modifying its problem description.

\newcommand{\dom}{\mathit{dom}}
\newcommand{\inn}{\mathit{in}}
\newcommand{\edge}{\mathit{edge}}
\newcommand{\vtx}{\mathit{vtx}}
\newcommand{\reach}{\mathit{reach}}
\newcommand{\distance}{\mathit{d}}
\newcommand{\minimize}{\mathit{minimize}}

Consider the Minimum Connected Dominating Set (MCDS) problem. A dominating set is a set of nodes such that every node in the graph is either in the set 
or has at least one neighbour in the set. The objective is to find a dominating set of minimum cardinality such that 
the subgraph induced by dominating nodes is connected. Let us look at the ASP encoding of the problem
\footnote{Based on the model from Potassco group in the second ASP competition: \url{http://dtai.cs.kuleuven.be/events/ASP-competition/encodings.shtml}}. 
A vertex $X$ is given in the input as
$\vtx(X)$, an edge from $X$ to $Y$ as $\edge(X,Y)$ and the edge relation is symmetric.

$$
\begin{array}{ll}
R_1 & \{ \dom(U) : \vtx(U) \}. \\

R_2 & \inn(V) \leftarrow \dom(V). \\
R_3 & \inn(V) \leftarrow \edge(U,V) \wedge \dom(U). \\
C_1 & \leftarrow \vtx(U) \wedge \neg \inn(U). \\

R_4 & \reach(U) \leftarrow \dom(U) \wedge \not\exists_{V < U} : \dom(V). \\
R_5 & \reach(V) \leftarrow \reach(U) \wedge \dom(V) \wedge \edge(U,V). \\
C_2 & \leftarrow \dom(U) \wedge \neg \reach(U). \\

O & \minimize \{\dom(U) : \vtx(U) \}.
\end{array}
$$

$R_1$ introduces a decision variable $\dom$ for each vertex specifying whether it is a dominating vertex or not.
$R_1$ is a \emph{choice rule} which means that for a given node $U$, this rule can be a justification for
$\dom(U)$ if it is true.
$R_2$, $R_3$ and $C_1$ model that every node must either be a dominating node or have
a neighbour that is dominating. This is done with the help of an auxiliary predicate $\inn$ that becomes true
when at least one of the conditions is met. $C_1$ says that there can be no node for which $\inn$ is false. 
$R_4$ and $R_5$ define the predicate $\reach$ that is used to model the connectivity constraint of the induced subgraph by the dominating nodes. 
$R_4$ encodes the base case for reachability, specifying that the node with the lowest index is reachable by definition. This choice is arbitrary
and its purpose can be satisfied by any criterion to select a dominating node. $R_5$ is a recursive case for reachability and it says that a dominating neighbour of
a reachable node is also reachable. The constraint $C_2$ says that all dominating nodes must be reachable. 
The objective, given by $O$, is to minimize the cardinality of the dominating set.


Let us modify MCDS such that the edges in the graph also have weights ($\edge(U,V,W)$ means that the edge from $U$ to $V$ has weight $W$) 
and there is an additional constraint that the diameter (maximum distance between any two nodes) of the dominating subgraph is less than a certain given value $K$.
This problem has applications in computer networks \cite{cds,cds_journal}.
Let $\distance(X,Y)$ represent the distance (shortest path) between two dominating nodes $X$ and $Y$.
In MCDS, it is sufficient to check for reachability of every dominating node from an arbitrary node to ensure connectedness. However, to enforce the new
constraint, we need a distance variable for each pair of nodes in the dominating set.
We can replace $R_4$, $R_5$ and $C_2$ in the above encoding of MCDS with the following:

$$
\begin{array}{ll}
R_4 & \distance(U,U) \leq 0 \leftarrow \dom(U). \\
R_5 & \distance(U,T) \leq \distance(V,T) + W \leftarrow \dom(T) \wedge \dom(U) \wedge \dom(V) \wedge \edge(U,V,W). \\
C_2 & \leftarrow \dom(U) \wedge \dom(V) \wedge \distance(U,V) > K.
\end{array}
$$

$R_4$ is the base case for $\distance$ and it says that the distance from a dominating node to itself is at most $0$.
$R_5$ is a recursive rule that specifies that for two dominating neighbours and a dominating node $T$, the distance
between one end of the node to $T$ is at most the distance between the other end and $T$, plus the weight of the edge.
Finally, the constraint $C_2$ establishes that the distance between any two dominating nodes must be at most $K$.
It is unnecessary to include the previous reachability rules since finite distances between all pairs of dominating
nodes implies that the dominating set is connected.

Rules like $R_4$ and $R_5$ on integer variables are clearly not supported by current ASP systems.
The semantics that we wish to associate with the distance variables is that firstly, if there are no rules supporting them,
then they are equal to $\infty$. Secondly, any rule for a distance variable justifies a value lower than $\infty$ and thirdly, the upper bounds
of these variables cannot form a circular justification. E.g. if there are two rules: $a \leq b$ and $b \leq a$, then any solution where
$a$ and $b$ are equal to a finite value should be rejected, and the only \emph{stable} solution should be one where both are equal to $\infty$.
The distance variable is essentially an \emph{upper-bound founded} (ub-founded) variable, for which the upper bound needs to be justified by some rule.
We can encode these upper-bound founded distance variables in ASP along with our desired semantics by replacing $R_4$, $R_5$, and $C_2$ as follows:
\newcommand{\dub}{\mathit{d_{ub}}}
$$
\begin{array}{l}
\dub(U,V,N) \leftarrow \dom(U) \wedge \dom(V) \wedge \dub(U,V,N-1), N < M. \\
\dub(U,U,0) \leftarrow \dom(U). \\
\dub(U,T,D+W) \leftarrow \dom(T) \wedge \dom(U) \wedge \dom(V) \wedge \edge(U,V,W) \wedge d(V,T,D). \\
d(U,V,D) \leftarrow \dom(U) \wedge \dom(V) \wedge \dub(U,V,D) \wedge \neg \dub(U,V,D-1). \\
\leftarrow \dom(U) \wedge \dom(V) \wedge \distance(U,V,D) \wedge D > K.
\end{array}
$$

In the above encoding, $M$ is a sufficiently large integer and $\dub(U,V,N)$ specifies that the distance between the dominating nodes $U$ and $V$ is \emph{at most}
$N$ (the subscript $ub$ stands for upper-bound). 
$d(U,V,D)$ is defined as the minimum value for which $\dub(U,V,D)$ is true. 
Unfortunately, an ASP solver on this encoding quickly runs into the grounding bottleneck problem as we increase edge weights and the bound on diameter.
This is the motivation of this work, i.e., to support founded numeric variables and rules like $R_4$ and $R_5$ without
grounding them.

The symmetric analog for a ub-founded variable is a \emph{lower-bound founded} (lb-founded) variable,
which is by default equal to $\minf$ ($\false$ for Boolean) and further rules for it can justify a greater value on its lower bound. 
In this generalization, ASP variables are simply lb-founded Boolean variables. 
For simplicity, we only consider lb-founded variables, and refer to them as founded variables. 
This simplification is possible because we can replace all ub-founded variables, their rules, and their constraints by corresponding lb-founded variables
with similar rules and constraints. E.g., for the above problem, let $d(U,V)$ represent the negative of the distance between $U$ and $V$,
then we can perform this transformation as follows:

$$
\begin{array}{ll}
R_4 & \distance(U,U) \geq 0 \leftarrow \dom(U). \\
R_5 & \distance(U,T) \geq \distance(V,T) - W \leftarrow \dom(T) \wedge \dom(U) \wedge \dom(V) \wedge \edge(U,V,W). \\
C_2 & \leftarrow \dom(U) \wedge \dom(V) \wedge \distance(U,V) < -K.
\end{array}
$$

An important point in the encoding of MCDS is that if we remove the reachability condition from the problem specification,
and $R_4$, $R_5$, and $C_2$ from the encoding, then the problem can be solved by a SAT solver just as efficiently as an ASP solver. Recall that
the only shortcoming of SAT solvers is related to modelling properties like reachability since the propositional semantics that they are based
on does not naturally model recursive definitions.
This leads us to the important observation that besides founded variables like $\reach$ and $\distance$, there can be variables in a problem
like the $\dom$ variables that are not founded.
Let us call them \emph{standard} variables, owing to the fact that these are the usual variables in CP solvers. Standard variables can be
assigned any value as long as all the constraints associated with them are satisfied. Founded variables, on the other hand,
need rules to define their values and without them, they are equal to some default value.

Since there are no rules and founded variables in CP and MIP solvers, MCDS with bounded diameter as defined above
cannot be efficiently solved by them. The MIP formulation as given in \cite{cds_journal} encodes each distance variable with $K$ propositional variables,
meaning that the problem size increases with $K$. Our encoding above that uses founded variables does not suffer from this problem.
This leads us to distinguish between the two types of grounding bottlenecks. One is caused in a system by the absence of its support for standard integer
variables. Let us call this type the \emph{standard grounding bottleneck}. The other type of grounding bottleneck
is caused by the lack of a system's capabilities to handle founded numeric variables, therefore, let us call it
\emph{founded grounding bottleneck}. 

In the next section, we formally define the semantics of \emph{Bound Founded Answer Set Programming} (BFASP), a formalism that generalizes
the stable model semantics to bounds over numeric variables, and allows for a richer set of rules for founded variables.
\footnote{The \minizinc encoding of MCDS with bounded diameter in BFASP is given Appendix A.}

\section{Bound Founded Answer Set Programming}
Let $\VV$ be the set of variables. We consider three types of variables: integer, real, and Boolean.
Furthermore, we divide the set of variables in two disjoint sets: standard $\ANV$ and \emph{founded} 
variables $\AFV$.
A \emph{domain} $D$ maps each variable $x \in \VV$ to a set of constant values $D(x)$. 
A \emph{valuation} (or assignment) $\theta$ over variables
$vars(\theta) \subseteq \VV$ maps each variable $x \in vars(\theta)$ to a value
$\theta(x)$.
A \emph{constraint} $c$ is a set of assignments over the variables $vars(c)$,
representing the solutions of the constraint. 
Given a constraint $c$, a variable $y \in \vars(c)$ is 
\emph{monotonically increasing (decreasing)}
in $c$ if for all solutions $\theta$ that satisfy $c$, increasing (decreasing) 
the value of $y$ also creates a solution, that is
$\theta'$ where $\theta'(y) > \theta(y)$, and $\theta'(x) = \theta(x), x \in
vars(c) - \{y\}$, is also a solution of $c$. 

A \emph{positive-CP} $P$ is a collection of constraints where each constraint is increasing in exactly one variable and decreasing in the rest. 
The \emph{minimal} solution of a positive-CP is an assignment $\theta$ that satisfies $P$ s.t. there is no
other assignment $\theta'$ that also satisfies $P$ and there exists 
a variable $v$ for which
$\theta'(v) < \theta(v)$. Note that for Booleans, $\true > \false$.
A satisfiable positive-CP $P$ always has a unique minimal solution. 
If we have bounds consistent propagators for all the constraints in the program, then the unique minimal solution can be found simply by 
performing bounds propagation on all constraints until a fixed point is reached, and then setting all variables to their lowest values.

A rule $r$ is a pair $(c,y)$ where $c$ is a constraint, $y \in \AFV$ is the 
head of the rule and it is increasing in $c$.
A bound founded answer set program (BFASP) $P$ is a tuple $(\ANV, \AFV, C, R)$ where $C$ and $R$ are sets of constraints
and rules respectively. Given a variable $y \in \AFV$, $\rules(y)$ is the set of rules with $y$ as their heads.

The reduct of a BFASP $P$ w.r.t. an assignment $\theta$, written $\red{P}{\theta}$, is a 
positive-CP made from each rule $r = (c,y)$ by replacing in $c$
each variable $x \in vars(c) - \{y\}$, if it is a standard variable or if $c$ is not decreasing in it,
by its value $\theta(x)$ to create a positive-CP constraint $c'$. Let $\red{r}{\theta}$ denote this constraint.
If $\red{r}{\theta}$ is not a tautology, 
it is included in the reduct.
An assignment $\theta$ is a stable model of $P$ iff i) it satisfies
all the constraints in $P$ and ii) it is the minimal model that satisfies $P^\theta$.

\begin{example}
Consider a BFASP with standard variable $s$, integer founded variables $a,b$, 
Boolean founded variables $x$ and $y$, and the rules:
$(a \geq 0, a)$, $(b \geq 0, b)$, $(a \geq b + s, a)$,
$(b \geq 8 \leftarrow x, b)$, $(x \leftarrow \neg y \wedge (a \geq 5), x)$.
Consider an assignment $\theta$ s.t. $\theta(x) = \true$, $\theta(y) = \false$, $\theta(b) = 8$,
$\theta(s) = 9$ and $\theta(a) = 17$. The reduct of $\theta$ is the positive-CP: $a \geq b + 9$, $b \geq 8 \leftarrow x$,
$x \leftarrow a \geq 5$. The minimal model that satisfies the reduct is equal to $\theta$, therefore, $\theta$ is a stable
model of the program. Consider another assignment $\theta'$ where all values are the same as in $\theta$, but
$\theta'(s) = 3$. Then, $P^{\theta'}$ is the positive-CP: $a \geq 0$, $b \geq 0$, $a \geq b + 3$, $b \geq 8 \leftarrow x$, $x \leftarrow a \geq 5$.
The minimal solution that satisfies this positive-CP is $M$ where $M(a) = 3$, $M(b) = 0$, $M(x) = M(y) = \false$.
Therefore, $\theta'$ is not a stable model of the program.
\end{example}

\section{Overview of the existing literature}
There are several approaches in the literature that aim at removing the standard grounding bottleneck from ASP systems.
A majority of these approaches work as follows: they introduce finite domain integer variables and constraints inside
the ASP program. The ASP solver passes these to a CP solver while maintaining a Boolean variable to
represent the truth value of each constraint that is in the ASP program. For a constraint $c$ that appears
in the program, this is done by \emph{reifying} the constraint, i.e., introducing a Boolean variable $b$ to represent whether the constraint is true. 
The ASP and CP solvers communicate using these introduced Boolean variables. E.g., if $b$
is set true by the ASP solver, then the constraint $c$ is enforced by the CP solver.
Since the ASP solver treats CP as a blackbox, it cannot directly learn clauses from the propagation performed
by the CP solver. Examples of systems that use this approach are 
the \printsolver{AC solver} algorithm \cite{mellarkod_ai}, \clingcon \cite{clingcon} and \ezcsp \cite{ezcsp}.
Recently, some systems have been introduced that overcome the limited learning by using a single solver that supports
both founded Booleans as well as standard integer variables and constraints over them.
One way to achieve this is to introduce standard integer variables inside an ASP solver, and extending ASP's propagation
engine to work like a CP solver \cite{constraintid,inca_ng}. The second approach is given in our earlier work \cite{inductive_cp}, and 
extends an existing CP solver with founded Boolean variables and normal rules. To implement the stable model semantics over these,
it implements the \emph{source pointer technique} \cite{smodels} to prune \emph{unfounded sets} \cite{vangelder} of variables
as a propagator, similar to the ASP solver \clasp \cite{clasp_journal}. 

\emph{Translating} in terms of its supported features a specification that is missing in a system is another way to remove 
standard grounding bottleneck. There are two approaches in the literature to accomplish this. The first approach provides a translation from an ASP program
augmented with numeric variables and constraints to a Mixed Integer Program \cite{asp_via_mip}. 
As discussed earlier, the non-recursive parts of the program are straight-forward to translate.
The non-trivial part is encoding rules that involve positive recursion. This is done using the \emph{level ranking}
mapping as given in \cite{lpsat}. The fundamental idea of the translation is that if there is an unfounded set in the solution of the original program,
then the mapping contains an inconsistent set of inequalities. 
The second approach \cite{translation_casp} encodes entire CP solving
into ASP using the well-known eager CP decompositions to SAT. Unfortunately, this \emph{a priori} translation of CP to SAT is already known to be highly
inefficient in the CP community where it is much more efficient to translate \emph{lazily} as in \emph{lazy clause generation} \cite{lcg}.

Compared to the standard grounding bottleneck, the focus on the founded grounding bottleneck has been relatively
weak. The formalism that is closest to BFASP in terms of removing this bottleneck is Fuzzy Answer Set Programming (FASP)
\cite{NieuwenborghCV06,blondeel2013fuzzy}.
The fuzzy atoms in FASP correspond to founded real variables in BFASP, and each logical connective in FASP can be
expressed as a rule form in BFASP. We have provided the translation in our previous work \cite{bfasp}, showing
that BFASP subsumes FASP. Most importantly, from the implementation point of view, the MIP based unfounded set detection algorithm 
\cite{JanssenHVC08} given for FASP only
detects unfounded sets in a complete solution, which means that it cannot prune partial solutions that contain unfounded sets.
Therefore, the algorithm has a similar shortcoming as \cmodels \cite{cmodels} has in case of Boolean unfounded sets.
Finally, lack of any good implementation for FASP makes it infeasible to carry out an empirical comparison of BFASP
and FASP. 


\section{Goals and current status of the research}
The broader goal of my PhD is to analyze the strengths and implementation techniques of ASP in order to enhance the existing
modelling and solving capabilities of constraint solvers. 
This overall goal can be divided into the following subgoals. 

\begin{itemize}
\item To define, study, implement, and evaluate BFASP in order to put it forward as a formalism, with an accompanying implementation, that does not suffer
from any kind of grounding bottleneck. This subgoal has been completed, and the most important features of BFASP were published in ICLP 2013 \cite{bfasp}. 
The paper defines the semantics of BFASP and presents an unfounded set algorithm that detects circular
sets of bounds and prunes them during search. It presents performance comparison of our implementation of BFASP with ASP on three benchmarks, and with CP on one benchmark,
and the results demonstrate the need for BFASP.

Prior to introducing BFASP, we extended a CP solver with founded Boolean variables and normal rules \cite{inductive_cp}. 
We implemented two known algorithms for unfounded set detection from the ASP literature inside the CP solver \chuffed 
and compared it with \clingcon on problems that involve inductive definitions as well as standard integer variables. 
\item To define and study the language of BFASP. 
As compared to ASP languages like \gringo that follow a very
restrictive grammar, the grammar for BFASP is very permissive and a user can write complex expressions as rules. 
Therefore, the first task in this subgoal is to simplify these rules to a small set of primitive rules. 
In other words, we want to extend the \emph{flattening} principles \cite{mznfn} used in constraint languages to BFASP.
Secondly, ASP grounders use \emph{bottom-up grounding} that generates as few \emph{useless rules} as possible. 
These are rules that can be removed without affecting the stable
solutions of the program. The second task is to generalize ASP bottom-up grounding technique for BFASP. Finally,
\emph{magic set rewriting} is a useful technique in logic programming that only instantiates rules
that are relevant to a given \emph{query}. Considering variables appearing in constraints and objective function comprise our
query, i.e. the set of variables whose values are of interest, the final task is to generalize the magic set transformation for BFASP. 
\item Identify research areas and benchmarks where BFASP can be applied and that cannot be efficiently solved by the current
ASP, CP, and Constraint ASP \cite{clingcon} systems. The doctoral programme could prove especially beneficial with regard to this subgoal.
\end{itemize}

\bibliographystyle{acmtrans}
\bibliography{paper}

\begin{thebibliography}{}

\bibitem[\protect\citeauthoryear{Aziz, Chu, and Stuckey}{Aziz
  et~al\mbox{.}}{2013}]{bfasp}
{\sc Aziz, R.~A.}, {\sc Chu, G.}, {\sc and} {\sc Stuckey, P.~J.} 2013.
\newblock Stable model semantics for founded bounds.
\newblock {\em Theory and Practice of Logic Programming\/}~{\em 13,\/}~4--5,
  517--532.
\newblock Proceedings of the 29th International Conference on Logic
  Programming.

\bibitem[\protect\citeauthoryear{Aziz, Stuckey, and Somogyi}{Aziz
  et~al\mbox{.}}{2013a}]{inductive_cp}
{\sc Aziz, R.~A.}, {\sc Stuckey, P.~J.}, {\sc and} {\sc Somogyi, Z.} 2013a.
\newblock Inductive definitions in constraint programming.
\newblock In {\em Proceedings of the Thirty-Sixth Australasian Computer Science
  Conference}, {B.~Thomas}, Ed. CRPIT, vol. 135. ACS, 41--50.

\bibitem[\protect\citeauthoryear{Balduccini}{Balduccini}{2009}]{ezcsp}
{\sc Balduccini, M.} 2009.
\newblock Representing constraint satisfaction problems in answer set
  programming.
\newblock In {\em ICLP09 Workshop on Answer Set Programming and Other Computing
  Paradigms (ASPOCP09)}.

\bibitem[\protect\citeauthoryear{Baral}{Baral}{2003}]{baral03}
{\sc Baral, C.} 2003.
\newblock {\em {Knowledge representation, reasoning and declarative problem
  solving}}.
\newblock Cambridge University Press.

\bibitem[\protect\citeauthoryear{Blondeel, Schockaert, Vermeir, and
  De~Cock}{Blondeel et~al\mbox{.}}{2013}]{blondeel2013fuzzy}
{\sc Blondeel, M.}, {\sc Schockaert, S.}, {\sc Vermeir, D.}, {\sc and} {\sc
  De~Cock, M.} 2013.
\newblock Fuzzy answer set programming: An introduction.
\newblock In {\em Soft Computing: State of the Art Theory and Novel
  Applications}. Springer, 209--222.

\bibitem[\protect\citeauthoryear{Buchanan, Sung, Boginski, and
  Butenko}{Buchanan et~al\mbox{.}}{2013}]{cds_journal}
{\sc Buchanan, A.}, {\sc Sung, J.~S.}, {\sc Boginski, V.}, {\sc and} {\sc
  Butenko, S.} 2013.
\newblock On connected dominating sets of restricted diameter.
\newblock {\em European Journal of Operational Research\/}.

\bibitem[\protect\citeauthoryear{de~Cat, Bogaerts, Devriendt, and
  Denecker}{de~Cat et~al\mbox{.}}{2013}]{constraintid}
{\sc de~Cat, B.}, {\sc Bogaerts, B.}, {\sc Devriendt, J.}, {\sc and} {\sc
  Denecker, M.} 2013.
\newblock Model expansion in the presence of function symbols using constraint
  programming.
\newblock In {\em {IEEE International Conference on Tools with Artificial
  Intelligence}}.

\bibitem[\protect\citeauthoryear{Drescher and Walsh}{Drescher and
  Walsh}{2010}]{translation_casp}
{\sc Drescher, C.} {\sc and} {\sc Walsh, T.} 2010.
\newblock A translational approach to constraint answer set solving.
\newblock {\em Theory and Practice of Logic Programming\/}~{\em 10,\/}~4-6,
  465--480.

\bibitem[\protect\citeauthoryear{Drescher and Walsh}{Drescher and
  Walsh}{2012}]{inca_ng}
{\sc Drescher, C.} {\sc and} {\sc Walsh, T.} 2012.
\newblock Answer set solving with lazy nogood neneration.
\newblock In {\em Technical Communications of the 28th International Conference
  on Logic Programming}. 188--200.

\bibitem[\protect\citeauthoryear{Gebser, Kaufmann, Neumann, and Schaub}{Gebser
  et~al\mbox{.}}{2007}]{clasp}
{\sc Gebser, M.}, {\sc Kaufmann, B.}, {\sc Neumann, A.}, {\sc and} {\sc Schaub,
  T.} 2007.
\newblock Conflict-driven answer set solving.
\newblock In {\em Proceedings of the 20th International Joint Conference on
  Artificial Intelligence}. MIT Press, 386.

\bibitem[\protect\citeauthoryear{Gebser, Kaufmann, and Schaub}{Gebser
  et~al\mbox{.}}{2012}]{clasp_journal}
{\sc Gebser, M.}, {\sc Kaufmann, B.}, {\sc and} {\sc Schaub, T.} 2012.
\newblock Conflict-driven answer set solving: From theory to practice.
\newblock {\em Artificial Intelligence\/}~{\em 187}, 52--89.

\bibitem[\protect\citeauthoryear{Gebser, Ostrowski, and Schaub}{Gebser
  et~al\mbox{.}}{2009}]{clingcon}
{\sc Gebser, M.}, {\sc Ostrowski, M.}, {\sc and} {\sc Schaub, T.} 2009.
\newblock Constraint answer set solving.
\newblock In {\em Proceedings of the 25th International Conference on Logic
  Programming}. Springer, 235--249.

\bibitem[\protect\citeauthoryear{Gelfond and Lifschitz}{Gelfond and
  Lifschitz}{1988}]{stable}
{\sc Gelfond, M.} {\sc and} {\sc Lifschitz, V.} 1988.
\newblock The stable model semantics for logic programming.
\newblock In {\em Proceedings of the Fifth International Conference on Logic
  Programming}. MIT Press, 1070--1080.

\bibitem[\protect\citeauthoryear{Janhunen}{Janhunen}{2004}]{lpsat}
{\sc Janhunen, T.} 2004.
\newblock Representing normal programs with clauses.
\newblock In {\em Proceedings of the 16th Eureopean Conference on Artificial
  Intelligence}. 358--362.

\bibitem[\protect\citeauthoryear{Janssen, Heymans, Vermeir, and Cock}{Janssen
  et~al\mbox{.}}{2008}]{JanssenHVC08}
{\sc Janssen, J.}, {\sc Heymans, S.}, {\sc Vermeir, D.}, {\sc and} {\sc Cock,
  M.~D.} 2008.
\newblock Compiling fuzzy answer set programs to fuzzy propositional theories.
\newblock In {\em Proceedings of the 24th International Conference on Logic
  Programming}. Springer Berlin Heidelberg, 362--376.

\bibitem[\protect\citeauthoryear{Kim, Wu, Li, Zou, and Du}{Kim
  et~al\mbox{.}}{2009}]{cds}
{\sc Kim, D.}, {\sc Wu, Y.}, {\sc Li, Y.}, {\sc Zou, F.}, {\sc and} {\sc Du,
  D.-Z.} 2009.
\newblock Constructing minimum connected dominating sets with bounded diameters
  in wireless networks.
\newblock {\em IEEE Transactions on Parallel and Distributed Systems\/},
  147--157.

\bibitem[\protect\citeauthoryear{Lierler and Maratea}{Lierler and
  Maratea}{2004}]{cmodels}
{\sc Lierler, Y.} {\sc and} {\sc Maratea, M.} 2004.
\newblock Cmodels-2: Sat-based answer set solver enhanced to non-tight
  programs.
\newblock In {\em LPNMR}. 346--350.

\bibitem[\protect\citeauthoryear{Liu, Janhunen, and Niemela}{Liu
  et~al\mbox{.}}{2012}]{asp_via_mip}
{\sc Liu, G.}, {\sc Janhunen, T.}, {\sc and} {\sc Niemela, I.} 2012.
\newblock Answer set programming via mixed integer programming.
\newblock In {\em Proceedings of the 13th International Conference on
  Principles of Knowledge Representation and Reasoning}. AAAI Press, 32--42.

\bibitem[\protect\citeauthoryear{Mellarkod, Gelfond, and Zhang}{Mellarkod
  et~al\mbox{.}}{2008}]{mellarkod_ai}
{\sc Mellarkod, V.~S.}, {\sc Gelfond, M.}, {\sc and} {\sc Zhang, Y.} 2008.
\newblock Integrating answer set programming and constraint logic programming.
\newblock {\em Annals of Mathematics and Artificial Intelligence\/}~{\em
  53,\/}~1-4, 251--287.

\bibitem[\protect\citeauthoryear{Mitchell}{Mitchell}{2005}]{sat_solver}
{\sc Mitchell, D.~G.} 2005.
\newblock A {SAT} solver primer.
\newblock {\em Bulletin of the EATCS\/}~{\em 85}, 112--132.

\bibitem[\protect\citeauthoryear{Nieuwenborgh, Cock, and Vermeir}{Nieuwenborgh
  et~al\mbox{.}}{2006}]{NieuwenborghCV06}
{\sc Nieuwenborgh, D.~V.}, {\sc Cock, M.~D.}, {\sc and} {\sc Vermeir, D.} 2006.
\newblock Fuzzy answer set programming.
\newblock In {\em Proceedings of Logics in Artificial Intelligence, 10th
  European Conference, JELIA 2006}. Springer Berlin Heidelberg, 359--372.

\bibitem[\protect\citeauthoryear{Ohrimenko, Stuckey, and Codish}{Ohrimenko
  et~al\mbox{.}}{2009}]{lcg}
{\sc Ohrimenko, O.}, {\sc Stuckey, P.~J.}, {\sc and} {\sc Codish, M.} 2009.
\newblock Propagation via lazy clause generation.
\newblock {\em Constraints\/}~{\em 14,\/}~3, 357--391.

\bibitem[\protect\citeauthoryear{Rossi, Beek, and Walsh}{Rossi
  et~al\mbox{.}}{2006}]{cp_handbook}
{\sc Rossi, F.}, {\sc Beek, P.~v.}, {\sc and} {\sc Walsh, T.} 2006.
\newblock {\em Handbook of Constraint Programming (Foundations of Artificial
  Intelligence)}.
\newblock Elsevier Science, New York, NY.

\bibitem[\protect\citeauthoryear{Simons, Niemel{\"a}, and Soininen}{Simons
  et~al\mbox{.}}{2002}]{smodels}
{\sc Simons, P.}, {\sc Niemel{\"a}, I.}, {\sc and} {\sc Soininen, T.} 2002.
\newblock Extending and implementing the stable model semantics.
\newblock {\em Artificial Intelligence\/}~{\em 138,\/}~1--2, 181--234.

\bibitem[\protect\citeauthoryear{Stuckey and Tack}{Stuckey and
  Tack}{2013}]{mznfn}
{\sc Stuckey, P.~J.} {\sc and} {\sc Tack, G.} 2013.
\newblock Minizinc with functions.
\newblock In {\em Proceedings of the 10th International Conference on
  Integration of Artificial Intelligence (AI) and Operations Research (OR)
  techniques in Constraint Programming}. Number 7874 in LNCS. Springer,
  268--283.

\bibitem[\protect\citeauthoryear{{Van Gelder}, Ross, and Schlipf}{{Van Gelder}
  et~al\mbox{.}}{1988}]{vangelder}
{\sc {Van Gelder}, A.}, {\sc Ross, K.~A.}, {\sc and} {\sc Schlipf, J.~S.} 1988.
\newblock Unfounded sets and well-founded semantics for general logic programs.
\newblock In {\em Proceedings of the ACM Symposium on Principles of Database
  Systems}. ACM, 221--230.

\end{thebibliography}

\pagebreak
\appendix
\section{\minizinc Encoding of Minimum Connected Dominating Set with Bounded Diameter}
\begin{verbatim}
int: N;                       %number of nodes
int: E;                       %number of edges
array[1..E] of 1..N: from;    %encodes an edge i (from[i], to[i])
array[1..E] of 1..N: to;
array[1..E] of int: weight;   %weight of an edge
int: K;                       %bound on diameter

array[1..N] of var bool: dom; %whether a node is dominating

% Dominating set constraint
constraint forall (n in 1..N) (
  dom[n] \/ exists(e in 1..E where from[e] = n) (dom[to[e]])
);

% Rules for negative distance
array[1..N,1..N] of var int: d :: founded;

rule forall (n in 1..N) (d[n,n] >= 0 :: head(d[n,n]));
rule forall (e in 1..E, n in 1..N) (
  d[from[e],n] >= d[to[e],n] - weight[e] 
    <- dom[from[e]] /\ dom[to[e]] /\ dom[n] :: head(d[from[e],n])
);

% Diameter constraint
constraint forall (u,v in 1..N where u != v) (
  dom[u] /\ dom[v] -> d[u,v] >= -K
);

% Objective to minimize the cardinality of dominating set
solve minimize sum (n in 1..N) (bool2int(dom[n]));

% A toy instance
N=4;
E=6;
K=35;
from   =[1, 2, 2, 3, 3, 4];
to     =[2, 1, 3, 2, 4, 3];
weight =[20,20,30,30,40,40];
\end{verbatim}
\end{document}